\documentclass[letterpaper, 10 pt, conference]{ieeeconf}  

\IEEEoverridecommandlockouts                              

\usepackage{amsmath,amsfonts}
\usepackage{algorithmicx}
\usepackage{algorithm,algpseudocode}
\usepackage{array}
\usepackage[caption=false,font=normalsize,labelfont=sf,textfont=sf]{subfig}
\usepackage{textcomp}
\usepackage{stfloats}
\usepackage{url}
\usepackage{verbatim}
\usepackage{graphicx}
\usepackage[percent]{overpic}
\usepackage{cite}
\usepackage[top=57pt, bottom=43pt, left=48pt, right=48pt]{geometry}
\usepackage[table]{xcolor}
\usepackage{booktabs}

\graphicspath{ {Images/} } 

\newcommand{\real}{\mbox{\rm I$\!$R}}

\newcommand{\bld}[1]{\mbox{\boldmath $#1$}} 

\usepackage{cuted}

\usepackage{xcolor}

\title{\LARGE \bf
Real-Time Nonlinear Model Predictive Control of Heavy-Duty Skid-Steered Mobile Platform for Trajectory Tracking Tasks
}

\author{Alvaro Paz, Pauli Mustalahti, Mohammad Dastranj* and Jouni Mattila
\thanks{\protect\rule{0.965\linewidth}{0.4pt} \indent This work was supported by the Business Finland partnership project ``Future All-Electric Rough Terrain Autonomous Mobile Manipulators'' (Grant \#2334/31/2022). Corresponding Author: Mohammad Dastranj.}
\thanks{All authors are with the Unit of Automation Technology and Mechanical Engineering, Faculty of Engineering and Natural Sciences, Tampere University, 33720 Tampere, Finland.
        {\tt\small \{mohammad.dastranj, jouni.mattila\}@tuni.fi}}}

\begin{document}

© 2025 IEEE. Personal use of this material is permitted.
Permission from IEEE must be obtained for all other uses,
including reprinting/republishing this material for advertising
or promotional purposes, collecting new collected works
for resale or redistribution to servers or lists, or reuse of
any copyrighted component of this work in other works.
This work has been submitted to the IEEE for possible
publication. Copyright may be transferred without notice,
after which this version may no longer be accessible.

\newpage

\maketitle
\thispagestyle{empty}
\pagestyle{empty}

\begin{abstract}
This paper presents a framework for real-time optimal controlling of a heavy-duty skid-steered mobile platform for trajectory tracking. The importance of accurate real-time performance of the controller lies in safety considerations of situations where the dynamic system under control is affected by uncertainties and disturbances, and the controller should compensate for such phenomena in order to provide stable performance. A multiple-shooting nonlinear model-predictive control framework is proposed in this paper. This framework benefits from suitable algorithm along with readings from various sensors for genuine real-time performance with extremely high accuracy. The controller is then tested for tracking different trajectories where it demonstrates highly desirable performance in terms of both speed and accuracy. This controller shows remarkable improvement when compared to existing nonlinear model-predictive controllers in the literature that were implemented on skid-steered mobile platforms. 
\end{abstract}
\section{INTRODUCTION}
Mobile robots are a common type of dynamic systems that facilitate tasks in various fields, and their control has been an active research topic in recent decades \cite{raj2022comprehensive}. While various types of mobile robots exist, the focus of this paper is on controlling a four-wheel skid-steered mobile platform. Use of four-wheels provides proper stability without the excess design complexity compared to using more wheels \cite{rubio2019review}, while the skid steering mechanism allows for better mobility on uneven terrain, simpler mechanical design and handling larger payloads \cite{khan2021comprehensive}. These benefits, while making this steering mechanism extremely popular for mobile robots, are significantly important for heavy-duty applications. 

\subsection{Background}
Optimization-based control approaches, especially model-predictive control (MPC), have been an attractive topic for research in recent years as they allow for achieving proper tracking performance with optimal control effort, increasing efficiency. There is a trade-off concerning the choice between linear or nonlinear MPC approaches. Linear MPC provides computationally light solutions while its performance is limited by the nonlinear nature of many physical systems. On the other hand, nonlinear MPC (NMPC) provides reliable performance over a wider set of applications, but at the cost of being computationally heavy, challenging for real-time implementations. Therefore, it is crucial to choose optimization algorithms for NMPC with the aim of real-time implementation \cite{diehl2009efficient}.

The multiple-shooting algorithm introduced by \cite{bock1984multiple} allows for creating an NMPC framework to provide reliable robust optimal control performance, while yielding real-time implementation \cite{diehl2006fast}. The real-time operation of the controller is of significant importance in the presence of effective uncertainties, including tire-ground interactions in the operation of heavy-duty mobile platforms. If the controller performance is much slower than the actual dynamics affecting the system's motion, the controller may not compensate for these effects and therefore, instability may occur.

\begin{figure}[t] 
	\centering
	\includegraphics[trim={0.0cm 0.0cm 0.0cm 0.0cm},clip,width=8.5cm]{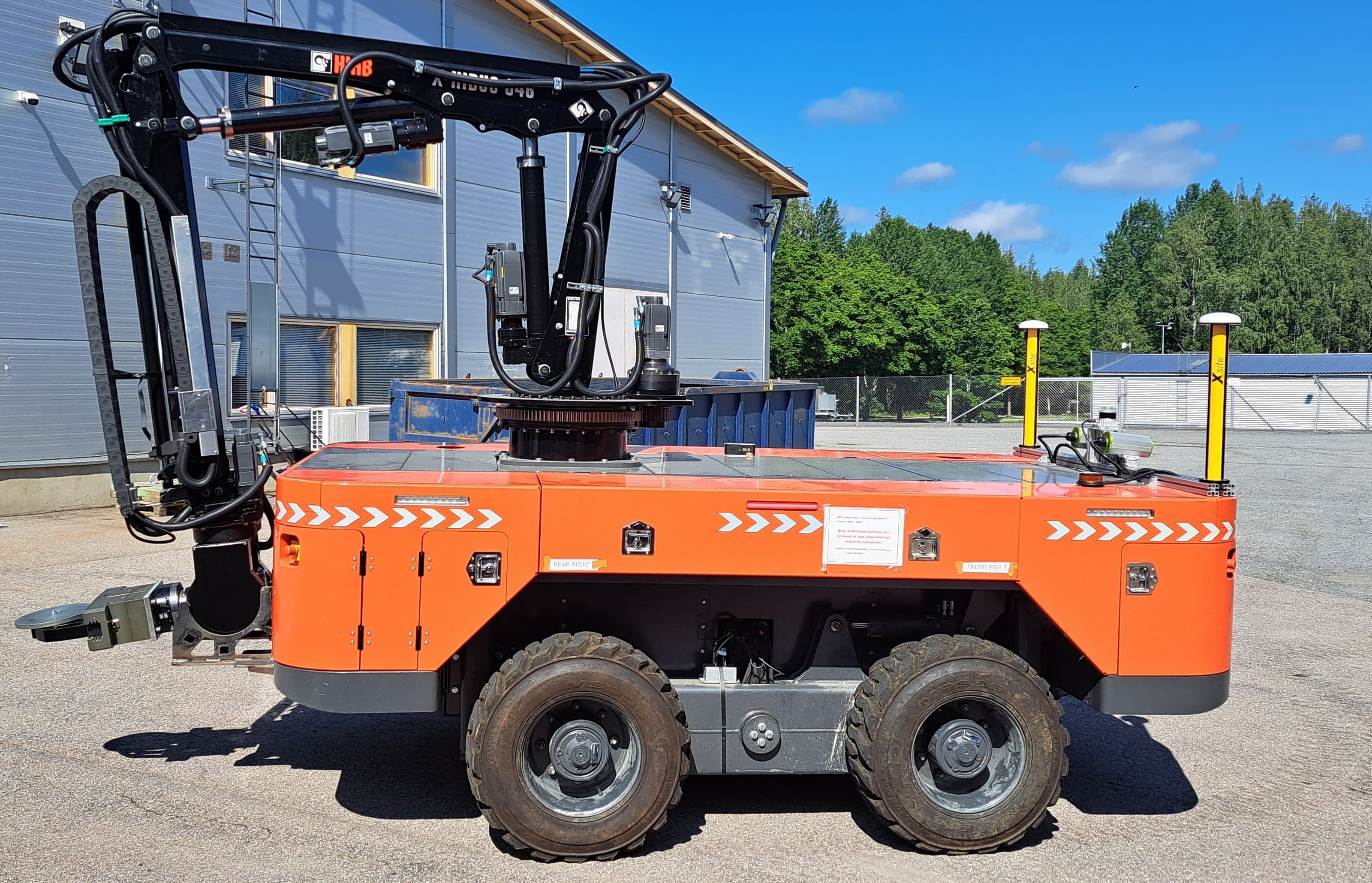}
	\caption{{\bf Skid-steered mobile platform.} This heavy-duty four-wheels robot is hydraulically driven and equipped with a stereo camera system for localization and wheel sensors for retrieving angular velocities.}
	\label{fig:platform}
\end{figure}

\subsection{Literature Review}
The efficient control with NMPC frameworks has demonstrated the high capabilities of NMPC for navigation tasks while including multiple sensors information and a wide variety of constraints and boundaries for legged robots \cite{fankhauser2018robust,grandia2023perceptive}, while the control of skid-steered mobile robots, using various control approaches, has been the topic of many pieces of research, it has been in recent years that NMPC was utilized for controlling this robot type. The applications of NMPC in trajectory tracking of skid-steered mobile robots has been reported in works like \cite{prado2020adaptive} where, by combining estimation features with NMPC, they achieved competitive real-time computations for S-shaped trajectories in a conventional-size platform. Furthermore, \cite{aro2023nonlinear} successfully included static obstacle avoidance in the NMPC for a small skid-steered platform with GPS localization. Also, for a better low-level controlling, \cite{wang2023trajectory} considered the actuator's dynamics in a heavy-duty robot demonstrating its efficiency when tracking circular trajectories. More complex trajectories on flat terrain, e.g. Lemniscate, are reported by \cite{aro2024robust} for a small mobile manipulator where its robust NMPC is endowed with passivity features.

Despite the proper implementation of these controllers in practical problems and the acceptable results, there is much room for improvement in terms of accuracy and speed. The accuracy is most important in scenarios such as confined or restricted work spaces, and the importance of computational speed is extremely noticeable in cases of fast dynamics. One remarkable risk regarding slow computation is the mismatch of estimations and the fast dynamics, which can cause imbalance in the system. Also, for the computational speed of the controller to be considered \textit{real-time}, it must be fast enough to perform the required operations between the two consecutive sampling steps of the required sensors of rather high frequency. High accuracy and real-time performance of the controller reduces the risks of damage to the personnel, the system, and/or the working environment. 

\subsection{Contributions}
This work is targeted to heavy-duty skid-steered platforms which their inherent features, e.g. bulky, heavy and slow motion, must be considered when designing a proper NMPC controller. We assume that our platform performs slow motion, which is reasonable for a 6,000 [$kg$] total weight robot with dimensions 3.7$\times$2.3 [$m$]; and considering that it is equipped with a heavy-duty manipulator on top of it, thus fast motions can easily compromise its dynamic equilibrium resulting in overturning events.

Then, we present a novel NMPC framework capable of tracking trajectories with high accuracy and real-time computation. Our NMPC is endowed with multiple features aiming at these two goals. We combine a multiple-shooting approach, that is fed with visual SLAM information, with wheel sensors' information for better accuracy. Additionally, we consider a three-fold approach with smooth-and-continuous step functions in the robot's dynamics for dealing with dead zones, which their effect is more considerable in heavy-duty robots due to the large inertia and friction. We also include a robust low-level controller for accurate tracking of the wheels' velocity.

In terms of fast computation, we achieve high performance by fulfilling the criteria of a proper real-time implementation. First, a combination of warm-start solutions, high-rate sensors sampling (1 kHz), and bounded maximum number of iterations in the solver result in a strategy to generate an optimal solution before the next sensor measurement arrives, i.e. execution time lower than the fastest sensor rate. Also, we implement algorithmic routines for managing the sensor's buffers. With this, we avoid unexpected data saturation and computational burden. These strategies endow our algorithm with deterministic computation time, which is another criterion for real-time implementations. 

To demonstrate the fast computation of optimized solutions, we perform a test with millions of online samples and prove that 98 [$\%$] of our NMPC samples are executed around 1 [$ms$], see Fig. \ref{fig:RT_sample}, which overcomes the times reported by \cite{prado2020adaptive}. Additionally, our time horizon is 3 times longer, i.e. $N$ = 30. We perform experiments with 3 trajectories, for which errors are reported in Table \ref{table:errors}. These are similar trajectories to the ones tested in \cite{prado2020adaptive,aro2023nonlinear,wang2023trajectory,aro2024robust} while they reported position errors around 1-5 [$m$] we obtained a maximum error of 6.2 [$cm$] demonstrating the high accuracy of our NMPC. Our maximum velocity error is 5.0 [$mm/s$]. The limitation in our work is that, even though the accuracy and computational speed reported in these works are improved in this research, the results of the previous works were obtained from tests in higher velocities while our platform only admits slow velocities.

The structure of the paper for following parts is as follows. In Section  \ref{sec:nlp}, the methodology used in this research is explained in parts. First, the dynamic system on which the controller is implemented is described. Then, the details of how the proposed solution of this research tackles the control objective are explained in detail. Section \ref{sec:nmpc} comprises of the details on how the NMPC works for the problem at hand, while the experimental results of implementing the proposed controller is presented in Section \ref{sec:results}. Finally, the analysis on how the proposed solution has handled the considered problem of real-time optimal trajectory tracking for a heavy-duty skid-steered mobile platform is concluded in Section \ref{sec:conclusion}.  

\section{Nonlinear Programming Problem}
\label{sec:nlp}
Let us define the transcription stage which is the process of converting an infinite optimal control problem (OCP) into a nonlinear programming problem (NLP) \cite{Bib:Betts}. This process plays an important role in the NMPC since it is the main algorithm running at each iteration of time, then it is crucial for achieving real-time performance and sensor synchronization.
\subsection{Skid-Steered Platform Model}
\label{sec:skid}
\begin{figure}[b] 
	\centering
	\includegraphics[trim={0.0cm 0.0cm 0.0cm 0.0cm},clip,width=7cm]{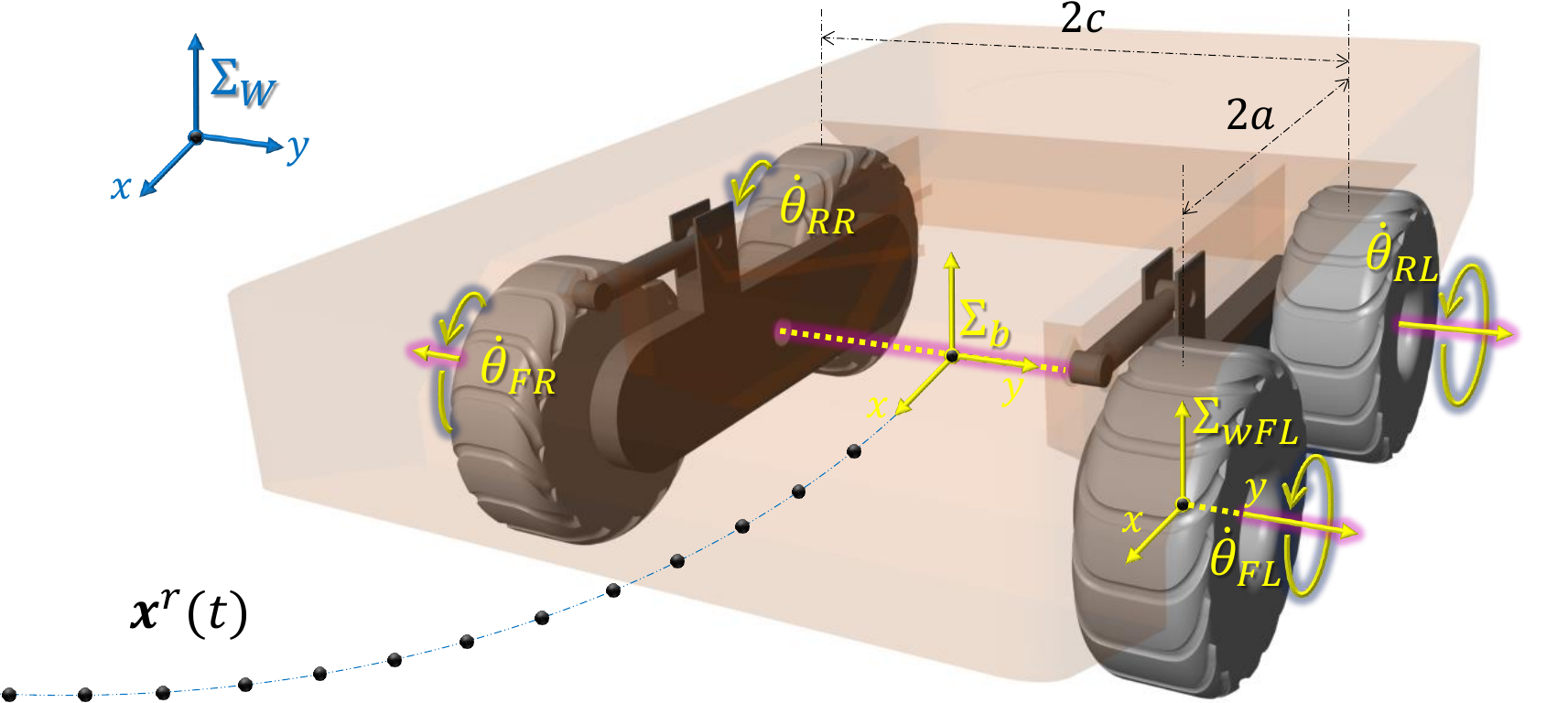}
	\caption{ Inertial global frame is denoted by $\Sigma_{W}$ and robot's mobile frame by $\Sigma_{b}$. The wheels' reference frames $\Sigma_{w\!F\!L}$, $\Sigma_{w\!F\!R}$, $\Sigma_{w\!R\!R}$ and $\Sigma_{w\!R\!L}$ are placed such that the wheels rotate around their $y$ axes of motion.}
	\label{fig:wheel}
\end{figure}

The wheeled mobile platform is depicted in Figs. \ref{fig:platform} and \ref{fig:wheel} where the four wheels are indexed by Front/Rear Right/Left and they have angular position $\theta$ and velocity $\dot{\theta}$.
The longitudinal and lateral distances between the wheels are $2a$ and $2c$, respectively. Thus, the robot's mobile reference frame $\Sigma_{b}$ is placed in the geometric center of the wheels' frames.

Let us adopt the skid steering assumption \cite{caracciolo1999trajectory} where right wheels have the same angular velocity $\dot{\theta}_{\!R}\!=\!\dot{\theta}_{\!F\!R}\!=\!\dot{\theta}_{\!R\!R}$, and similarly for the left wheels with velocity $\dot{\theta}_{\!L}\!=\!\dot{\theta}_{\!F\!L}\!=\!\dot{\theta}_{\!R\!L}$. Then, by neglecting the longitudinal slip of the wheels, the longitudinal linear velocity of each wheel frame is computed as $v_w \!=\! r\dot{\theta}$ where $r$ is the effective rolling radius of each wheel, assumed to be the same for all wheels. If we combine the wheel frame velocities, then the linear velocity $v_x$ of frame $\Sigma_{b}$, in its $x$ axis, and its angular velocity $\omega_z$, around its $z$ axis, expressed in the local frame are retrieved from the first order kinematic map \cite{kozlowski2004modeling}
\begin{equation}
    \bld{\nu}^{(b)} \ = \ \bld{J}\,\dot{\!\bld{\theta}}
\end{equation}
where $\bld{J}\in\real^{3\times2}$ is the Jacobian matrix
\begin{equation}
    \bld{J} \ = \ \dfrac{r}{2} \begin{bmatrix}
        1 & 0 & 1/c \\
        1 & 0 & -1/c
    \end{bmatrix}^{\top},
\end{equation}
the vector $\,\dot{\!\bld{\theta}} = [ \dot{\theta}_{\!R} \ \dot{\theta}_{\!L}]^{\top}\!\in\!\real^{2}$ contains the wheels' velocities and $\bld{\nu}^{(b)}$ is the twist, expressed in $\Sigma_{b}$, with vector form
\begin{equation}
    \bld{\nu}^{(b)} \ = \ \begin{bmatrix}
        v_x & v_y & \omega_z
    \end{bmatrix}^{\top}
\end{equation}
where $v_y=0$ if we assume the frame $\Sigma_{b}$ always moves tangent to the path.

Let us now assume a flat terrain and adopt a screw theory formulation where the Lie group $SE(2)= \real^{2} \!\times\! SO(2)$ describes all possible motions among two reference frames. For instance, $\bld{G}_{\!W}^{b} \in SE(2)$ encapsulates the position $\bld{p}_{\!W}^{b} \in \real^2$ and orientation $\alpha$ of the reference frame $\Sigma_{b}$ with respect to frame $\Sigma_{W}$ with the next matrix and vector representations
\begin{equation}
    \bld{G} \ = \ \begin{bmatrix}
        \bld{R} & \bld{p} \\
        \bld{0} & 1
    \end{bmatrix} \in \real^{3\times3} \quad \text{and} \quad \bld{x} \ = \ \begin{bmatrix}
         \bld{p} \\
         \alpha
    \end{bmatrix} \in \real^{3}
\end{equation}
where $\bld{R} \in SO(2)$ is the rotation matrix parameterized by $\alpha$ and $SO(2)$ is the Lie group for orthogonal matrices.

Since the twist $\bld{\nu}^{(b)}$ is an element of the Lie algebra $se(2)$, which is the algebra associated with $SE(2)$, then it can be transformed to the inertial frame $\Sigma_W$ by means of the adjoint mapping $\mbox{Ad}_{G}:SE(2)\rightarrow \real^{3\times3}$ as follows \cite{Bib:murray}
\begin{equation}
    \bld{\nu}^{(W)} \ = \ \mbox{Ad}_{G_{\!_{W}}^{_{b}}} \bld{\nu}^{(b)} \ = \ \mbox{Ad}_{G_{\!_{W}}^{_{b}}} \bld{J}\,\dot{\!\bld{\theta}}
    \label{eq:fsd}
\end{equation}
where $\bld{\nu}^{(W)}$ is the twist, which is assumed to be the time derivative of $\bld{x}$, i.e. $\bld{\nu}^{(W)} = \dot{\bld{x}}$, then (\ref{eq:fsd}) can be expressed by the nonlinear continuous-time model
\begin{equation}
    \dot{\bld{x}}(t) \ = \ \bld{f}( \bld{x}(t), \ \dot{\!\bld{\theta}}(t) )
    \label{eq:sys}
\end{equation}
since $\bld{x}$ and $\bld{G}_{\!_{W}}^{_{b}}$ are different representations of the same element.
\subsection{Transcription Procedure}
\label{sec:sklcvfus}
Let $\bld{x}^{r}(t)$ be a continuous-and-differentiable time reference trajectory on the manifold $SE(2)$ and let $\dot{\bld{x}}^{r}(t)$ be its first time derivative, then we can define the following constrained optimal control problem for tracking $\bld{x}^{r}(t)$ and considering $\dot{\!\bld{\theta}} (t)$ as our control variable 
\begin{eqnarray}
\hspace*{-0.65cm}\underset{\dot{\!\bld{\theta}} (t)}{\operatorname{min}} & \!\!\! & \tfrac{1}{2} \int_{0}^{T} \dot{\!\bld{\theta}}(t)^{\!\top} \ \dot{\!\bld{\theta}}(t) \ dt
\label{eq:costf1} \\
\nonumber \\
\hspace*{-0.65cm} \mbox{subject to} & \!\!\! &
\left\{\begin{array}{lcl}
\dot{\bld{x}}(t) & = & \bld{f}( \bld{x}(t), \ \dot{\!\bld{\theta}}(t) ) \vspace*{-0.0cm} \\
\bld{x}(t) & = & \bld{x}^r (t) \vspace*{-0.0cm} \\
\dot{\bld{x}}(t) & = & \dot{\bld{x}}^r (t) \vspace*{-0.0cm} \\
\dot{\bld{x}}_{{\scriptsize \mbox{min}}} & \leq & \dot{\bld{x}}(t) \ \ \leq \ \ \ \dot{\bld{x}}_{{\scriptsize \mbox{max}}} \vspace*{-0.0cm} \\
\dot{\!\bld{\theta}}_{{\scriptsize \mbox{min}}} & \leq & \ \dot{\!\bld{\theta}}(t) \ \ \leq \ \ \ \dot{\!\bld{\theta}}_{{\scriptsize \mbox{max}}} \vspace*{-0.0cm} \\
\end{array} \right. \ ,
\label{eq:const1}
\end{eqnarray}
where the cost function minimizes the control over the time $t\in[0,T]$. The constraints stand for the system dynamics (\ref{eq:sys}), the reference trajectory tracking with the system state $\bld{x}(t)$ and state-control boundaries where $\dot{\bld{x}}_{{\scriptsize \mbox{min}}}$ and $\dot{\!\bld{\theta}}_{{\scriptsize \mbox{min}}}$ are the lower limits for the platform velocity and wheels' angular velocity while $\dot{\bld{x}}_{{\scriptsize \mbox{max}}}$ and $\dot{\!\bld{\theta}}_{{\scriptsize \mbox{max}}}$ are their upper limits.

In order to transcribe the aforementioned OCP into a NLP that describes our NMPC with time horizon $N\in \mathbb{N}$, we first discretize the time as $t_k \, \forall \, k\in\{1 \!\cdots\! N\}$, then the time increment is ${\Delta_t} = t_{k+1}-t_k$. Since the state and control at time $k$ are referred to as $\bld{x}_k$ and $\dot{\!\bld{\theta}}_k$, then the system (\ref{eq:sys}) is evaluated as $\dot{\bld{x}}_{k}=\bld{f}( \bld{x}_k, \ \dot{\!\bld{\theta}}_k )$ and can be time integrated as
\begin{equation}
    \hat{\bld{x}}_{k+1} \ = \ \text{integrator}( \bld{x}_{k}, \dot{\bld{x}}_{k}, \Delta_t )
    \label{eq:int}
\end{equation}
where the $\text{integrator}( \cdot )$ function can perform numeric integration techniques, e.g. Euler, Runge-Kutta, etc. for approximating the state at $k\!+\!1$ as $\hat{\bld{x}}_{k+1}$.

By rolling out the discrete states and controls over the time horizon, we can define the following decision variable
\begin{equation}
    \bld{z} \ = \ [ \ \bld{x}_{0}^{\top} \ \,\dot{\!\bld{\theta}}_{0}^{\top} \ \cdots \ \bld{x}_{k}^{\top} \ \,\dot{\!\bld{\theta}}_{k}^{\top} \ \cdots \ \bld{x}_{\!N\!-\!1\!}^{\top} \ \,\dot{\!\bld{\theta}}_{\!N\!-\!1\!}^{\top} \ \bld{x}_{N}^{\top} \ ]^{\top} \in\real^{5N\!+3} ,\nonumber
\end{equation}
then at times $t_k$ and $t_N$ we can define a running cost $L_k (\cdot)$ and a terminal cost $L_N (\cdot)$ with multi-objective functions
\begin{eqnarray}
    L_k ( \bld{x}_{k}, \bld{x}_{k}^{r}, \,\dot{\!\bld{\theta}}_{k} ) & \!\!\!=\!\!\! & \left\| \bld{x}_{k} \!-\! \bld{x}_{k}^{r} \right\|_{Q_x}^2 + \left\| \dot{\bld{x}}_{k} \!-\! \dot{\bld{x}}_{k}^{r} \right\|_{Q_{\dot{x}}}^2 + || \,\dot{\!\bld{\theta}}_{k} ||_R^2 \nonumber \\
    L_{\!N} ( \bld{x}_{\!N}, \bld{x}_{\!N}^{r} ) & \!\!\!=\!\!\! & \left\| \bld{x}_{N} \!-\! \bld{x}_{N}^{r} \right\|_{Q_{x\!N}}^2 + \left\| \dot{\bld{x}}_{N} \!-\! \dot{\bld{x}}_{N}^{r} \right\|_{Q_{\dot{x}\!N}}^2 \nonumber
\end{eqnarray}
where $\bld{x}_{k}^{r}$ is the reference trajectory at time $t_k$ and $\dot{\bld{x}}_{k}^{r}$ is its time derivative. The objectives apply the weighted quadratic norm $|| \cdot ||^2$ where $R\in\real^{2\times 2}$ is the weighting matrix that minimizes the input control and the matrices $Q_x , Q_{\dot{x}}\in\real^{3\times 3}$ penalize the tracking deviation in position and velocity. Similarly, the weighting matrices $Q_{x\!N} , Q_{\dot{x}\!N}\in\real^{3\times 3}$ penalize the same deviation at $t_N$.

A state feedback controller can thus be defined by the following optimization problem considering the current ($t_0$) state and control variable measurements as $\bld{x}_{{\scriptsize \mbox{msr}}}$ and $\,\dot{\!\bld{\theta}}_{{\scriptsize \mbox{msr}}}$
\begin{equation}
	\hspace*{-0.17cm}\underset{\bld{z}}{\text{arg min}}  \ \ \ J \ = \ \tfrac{1}{2} \sum_{k=1}^{N-1} \hspace{-0.05cm} L_k ( \bld{x}_{k}, \bld{x}_{k}^{r}, \,\dot{\!\bld{\theta}}_{k} ) + L_{\!N} ( \bld{x}_{\!N}, \bld{x}_{\!N}^{r} )
	\label{eq:costf2}
\end{equation}
\hspace{0.0cm}subject to
\begin{subequations}
\begin{eqnarray}
\bld{x}_{0} & = & \bld{x}_{{\scriptsize \mbox{msr}}} \ \qquad \text{and} \ \qquad \,\dot{\!\bld{\theta}}_{0} \ = \ \,\dot{\!\bld{\theta}}_{{\scriptsize \mbox{msr}}} \label{eq:res_a} \\
\bld{x}_{k+1} & = & \hat{\bld{x}}_{k+1} \label{eq:res_b} \\
\bld{x}_{{\scriptsize \mbox{min}}} & \leq & \bld{x}_{k} \ \ \leq \ \ \ \bld{x}_{{\scriptsize \mbox{max}}} \qquad k=0 \cdots N \label{eq:res_c} \\
\dot{\!\bld{\theta}}_{{\scriptsize \mbox{min}}} & \leq & \ \dot{\!\bld{\theta}}_k \ \ \leq \ \ \ \dot{\!\bld{\theta}}_{{\scriptsize \mbox{max}}} \qquad k=0 \cdots N\!\!-\!\!1 \label{eq:res_d} \\
\ddot{\!\bld{\theta}}_{{\scriptsize \mbox{min}}} & \leq & \ \ddot{\!\bld{\theta}}_k \ \ \leq \ \ \ \ddot{\!\bld{\theta}}_{{\scriptsize \mbox{max}}} \qquad k=0 \cdots N\!\!-\!\!1 \label{eq:res_e}
\end{eqnarray}
\label{eq:costf2_}
\end{subequations}
where $J$ is the objective function that combines running and terminal costs. The constraints (\ref{eq:res_a}) force the initial state and control to be the current measurements to achieve a closed-loop feedback. Restriction (\ref{eq:res_b}) forces the decision-variable states to be the same as the ones approximated by (\ref{eq:int}), which is what actually creates the multiple-shooting matching. The constraints (\ref{eq:res_c}), (\ref{eq:res_d}) and (\ref{eq:res_e}) bound the system states $\bld{x}_{k}$, the controls $\dot{\!\bld{\theta}}_k$ and the time derivative of controls $\ddot{\!\bld{\theta}}_k$ through minimum limits such as $\bld{x}_{{\scriptsize \mbox{min}}}$, $\dot{\!\bld{\theta}}_{{\scriptsize \mbox{min}}}$ and $\ddot{\!\bld{\theta}}_{{\scriptsize \mbox{min}}}$, and maximum limits $\bld{x}_{{\scriptsize \mbox{max}}}$, $\dot{\!\bld{\theta}}_{{\scriptsize \mbox{max}}}$ and $\ddot{\!\bld{\theta}}_{{\scriptsize \mbox{max}}}$. The wheels' angular acceleration $\ddot{\!\bld{\theta}}_{k}$ can be retrieved through numeric differentiation since the controls $\dot{\!\bld{\theta}}_{k}$, $\dot{\!\bld{\theta}}_{k+1}$ and the increment of time $\Delta_t$ are known.

\subsection{Dealing with Dead Zones}
\label{sec:fdjhk}
\begin{figure}[t] 
	\centering
	\includegraphics[trim={2.4cm 0.15cm 3.7cm 0.5cm},clip,width=8.5cm]{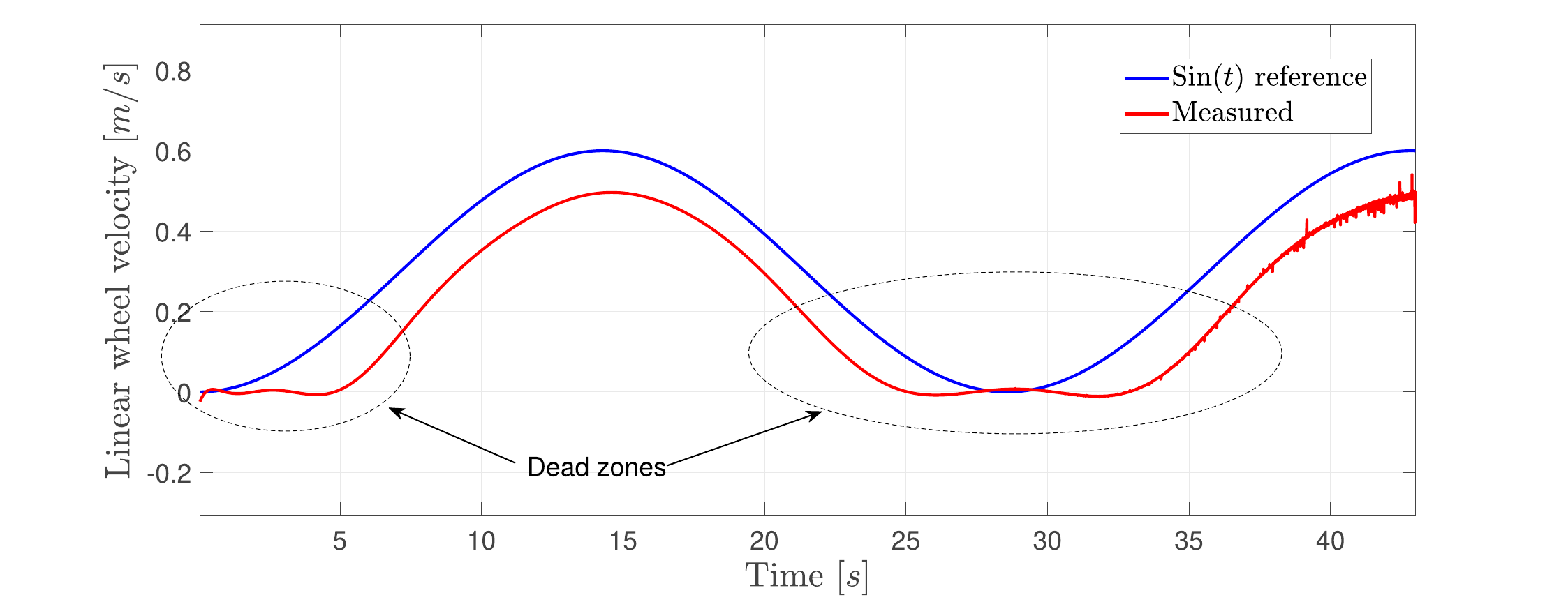}
	\caption{{\bf Open-loop test of linear wheel velocities.} We command a sine function to the platform and detected marginal performance due to death zones. The measured signal varies only up to reaching a threshold in the reference. The tracking error shows the need of a robust low level controller. }
	\label{fig:death_zone}
\end{figure}

For certain minimum values of the input control $\dot{\!\bld{\theta}}_k^*$, no actuator response is observed. This well-known dead-zone effect is caused by the inertia and friction of the mechanism; therefore, it is more significant in heavy-duty platforms. We perform an open-loop test in our 6,000 [$kg$] platform, see Fig. \ref{fig:death_zone}, to visualize this effect in our low-level controller and actuators \cite{shahna2025anti}.

In order to mitigate the dead-zone effect, and its induced mechanical chattering, we combine a three-fold approach. First, minimizing the control input through the criteria $|| \,\dot{\!\bld{\theta}}_{k} ||_R^2$ to reduce the chattering and constrain it with minimum and maximum values for avoiding close-to-zero values. Second, including the dead-zone effect into the system model (\ref{eq:sys}) by means of a smooth-and-continuous approximation step function \cite{khalil2002nonlinear} which produces no system evolution for values of $\dot{\!\bld{\theta}}_k$ below a threshold. Third, the aforementioned treatment of the dead zones can cause a non-smooth transition among two consecutive control inputs $\dot{\!\bld{\theta}}_k$ and $\dot{\!\bld{\theta}}_{k+1}$. This, directly increases the chattering of the platform; then, the variation in time of this two values is bounded by means of the acceleration constraint (\ref{eq:res_e}).

\section{Real-Time Nonlinear Model Predictive Control}
\label{sec:nmpc}
\begin{figure}[b] 
	\centering
	\includegraphics[trim={0.0cm 0.0cm 0.0cm 0.0cm},clip,width=8.6cm]{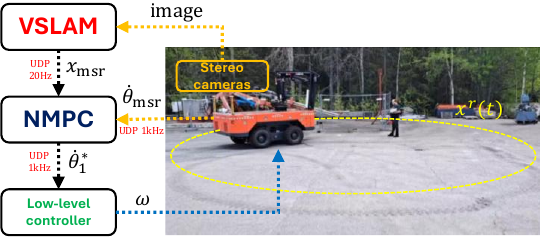}
	\caption{ {\bf Experimental setup.} The VSLAM system streams the robot pose $\bld{x}_{{\scriptsize \mbox{msr}}}$ at 20 Hz to the NMPC block while wheels' angular velocity sensors stream $\,\dot{\!\bld{\theta}}_{{\scriptsize \mbox{msr}}}$ at 1 kHz. Our NMPC computes an optimal control $\dot{\!\bld{\theta}}^{*}$ every 1 ms to the low-level controller which commands hydraulic motor's RPM $\omega$ runs in Beckoff system. All computation is performed on board.}
	\label{fig:overview}
\end{figure}

The experimental setup is depicted in Fig. \ref{fig:overview} where both sensors, VSLAM and wheels' angular velocity, stream at 20Hz and 1kHz, respectively. In order to achieve real-time controlling, we implement a set of techniques that enable our NMPC (\textbf{Algorithm \ref{algo:nmpc}}) for running in competitive times.

For instance, line 3 of the \textbf{Algorithm \ref{algo:nmpc}} generates an optimized code for the NLP and its first partial derivatives. Such code is integrated into a C++ numeric nonlinear solver where the combination of symbolic functions, BFGS \cite{Bib:Nocedal} and warm start generate an NMPC capable of finding an optimal solution in less time than the sensors' sampling rates, which is crucial for the real-time aspect where the computation time must be less than the sampling time. From lines 5 to 8, the solver prepares a suitable warm start solution by allowing a small tolerance in the solver and high number of iterations for getting a refined solution. This solution is then used as the primal solution of the online routines (lines 10-17). According to the requirements for real time, we allow a fixed small number of solver iterations so the computation time is bounded, then the NLP solution is improved by the high frequency of the NMPC iterations (lines 10-17) and not by convergence iterations of the solver at each NMPC step (line 14) \cite{grandia2023perceptive}. Additionally, real-time schemes must avoid overflows, burdens and unexpected latencies in communication, thus we implement smart pointers and the UDP communication protocol to connect the NMPC block where the buffers are managed in an efficient way to avoid queues and delaying behaviors which can affect the performance of the pipeline flow.
	\begin{algorithm}[t]
		\scriptsize{
			\begin{algorithmic}[1]
            \State \textbf{Offline (one-time setup):}
				\State \mbox{Define} $N$, $\bld{x}^{r}\!(t)$, $Q$, $R$, $\bld{x}_{{\scriptsize \mbox{min}}}$, $\dot{\!\bld{\theta}}_{{\scriptsize \mbox{min}}}$, $\ddot{\!\bld{\theta}}_{{\scriptsize \mbox{min}}}$, $\bld{x}_{{\scriptsize \mbox{max}}}$, $\dot{\!\bld{\theta}}_{{\scriptsize \mbox{max}}}$, $\ddot{\!\bld{\theta}}_{{\scriptsize \mbox{max}}}$
                \State C++ symbolic code generation of the NLP described in \ref{sec:nlp}.
                \State Integrate code into a numeric solver for a real-time execution.
                \State \textbf{Initialization (preparing warm-start solution):}
                \State Receive initial $\bld{x}_{{\scriptsize \mbox{msr}}}$ and $\,\dot{\!\bld{\theta}}_{{\scriptsize \mbox{msr}}}$ from sensors via \texttt{UDP}.
                \State Compute reference trajectory $\{\bld{x}^{r}_{k}\}_{k=0}^{N}$
                \State Call solver with small tolerance and allowing high number of iterations.
                \State \textbf{Online RT NMPC:}
				\While{NMPC is On}
                \State Read current measurements $\bld{x}_{{\scriptsize \mbox{msr}}}$ and $\,\dot{\!\bld{\theta}}_{{\scriptsize \mbox{msr}}}$ from sensors via \texttt{UDP}.
                \State Update reference $\{\bld{x}^{r}_{k}\}_{k=0}^{N}$ where $t_0$ is the current time.
                \State Warm-start solver using shifted previous solution as primal solution.
                \State Solve NLP with fixed-and-small number of iterations.
                \State Extract and stream optimal control $\dot{\!\bld{\theta}}^{*}_{1}$ via \texttt{UDP}.
                \State Shift optimal control sequence $\dot{\!\bld{\theta}}^{*}_{k}$ for warm-starting the next iteration.
                \EndWhile
			\end{algorithmic}
		}
		\caption{\small{{\textbf {Real-Time NMPC Execution Pipeline}}}}
		\label{algo:nmpc}
	\end{algorithm}
\section{Experimental Results}
\label{sec:results}
All experiments were performed on an industrial-grade computing platform (Nuvo-9160GC) equipped with an Intel Core i9 processor, an NVIDIA RTX 3050 GPU, 32 GB of RAM, and a 1 TB SSD, running a distribution of Linux operating system with C++ implementation. This is mounted on the mobile platform and connected through Ethernet. The NLP is coded in M{\footnotesize ATLAB} using C{\footnotesize ASADI} for C++ code generation. The NLP and its first order symbolic derivatives are linked to the nonlinear solver I{\footnotesize POPT}. The library B{\footnotesize OOST} is included for managing the UDP connections. The low-level control is running on a Beckhoff CX2043 industrial PC with a 1 [$ms$] sampling rate. We have used ORB-SLAM for localization as implemented in \cite{mur2017orb}.
\begin{figure}[t] 
	\centering
	\includegraphics[trim={0.0cm 0.0cm 0.0cm 0.0cm},clip,width=5cm]{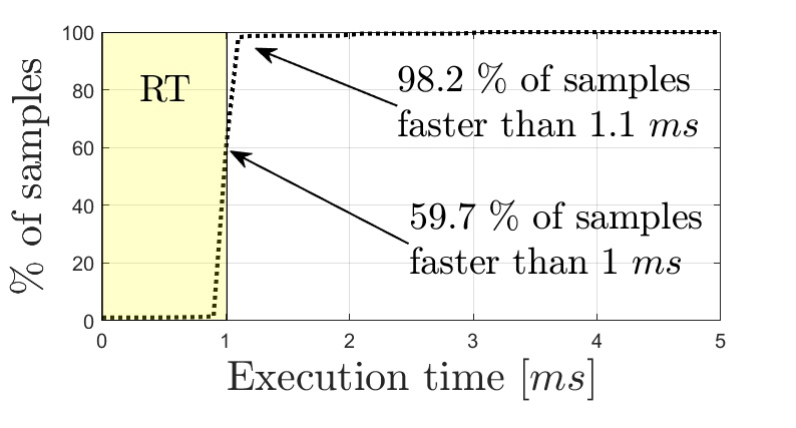}
	\caption{From the following experiments, one million samples of the online NMPC are analyzed and showing the percentage of samples faster than certain execution time. These samples include the whole pipeline depicted in Fig. \ref{fig:overview} then the NMPC-block execution time can be smaller in reality.}
	\label{fig:RT_sample}
\end{figure}

In order to achieve real-time execution time for the following experimentation, see Fig. \ref{fig:RT_sample}, we set $N=30$, $\Delta_t = 0.1$ and use Euler step as the numeric integrator. I{\footnotesize POPT} solver 
is enabled to use BFGS for Hessian approximation, its tolerance is set to $1E-6$ and the maximum number of iterations as $1$. This last setting allows for performing only one iteration of SQP by NMPC iteration where the NLP solution is optimized due to the warm starting and high rate of sensors sampling \cite{grandia2023perceptive}. The weighting matrices are set constant as $R\!=\!\text{diag}(0.2,0.2)$ and $Q_x \!=\! Q_{x\!N} \!=\! \text{diag}(20,20,12)$. Also, the limits are set constant as $\bld{x}_{{\scriptsize \mbox{min}}} \!=\! -[\text{Inf}, \text{Inf}, \text{Inf}]$, $\bld{x}_{{\scriptsize \mbox{max}}} \!=\! [\text{Inf}, \text{Inf}, \text{Inf}]$, $\dot{\!\bld{\theta}}_{{\scriptsize \mbox{min}}} \!=\! [0.1, 0.1]$, $\dot{\!\bld{\theta}}_{{\scriptsize \mbox{max}}} \!=\! [0.8, 0.8]$, $\ddot{\!\bld{\theta}}_{{\scriptsize \mbox{min}}} \!=\! -[0.2, 0.2]$ and $\ddot{\!\bld{\theta}}_{{\scriptsize \mbox{max}}} \!=\! [0.2, 0.2]$. Please note that the positive values in $\dot{\!\bld{\theta}}_{{\scriptsize \mbox{min}}}$ help the system to avoid the dead zones by assuming that the platform performs only forward motion in the following trajectories.

\begin{figure}[t]
\centering
\begin{overpic}[width=5.0cm,trim={9.8cm 0.4cm 10.0cm 1.4cm},clip]{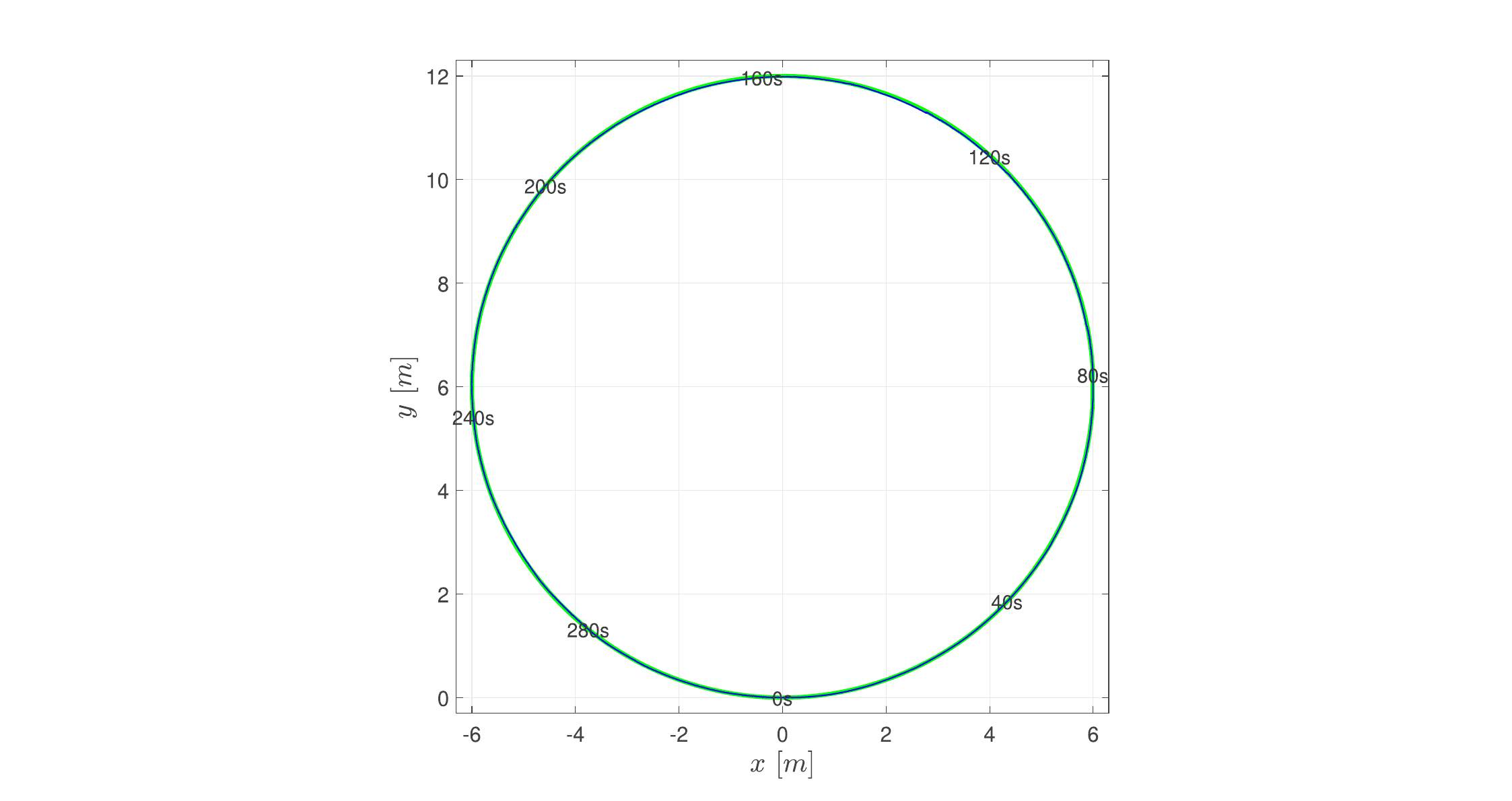}
    \put(42,47){\includegraphics[width=2cm,trim={9.3cm 2.0cm 10.8cm 1.4cm},clip]{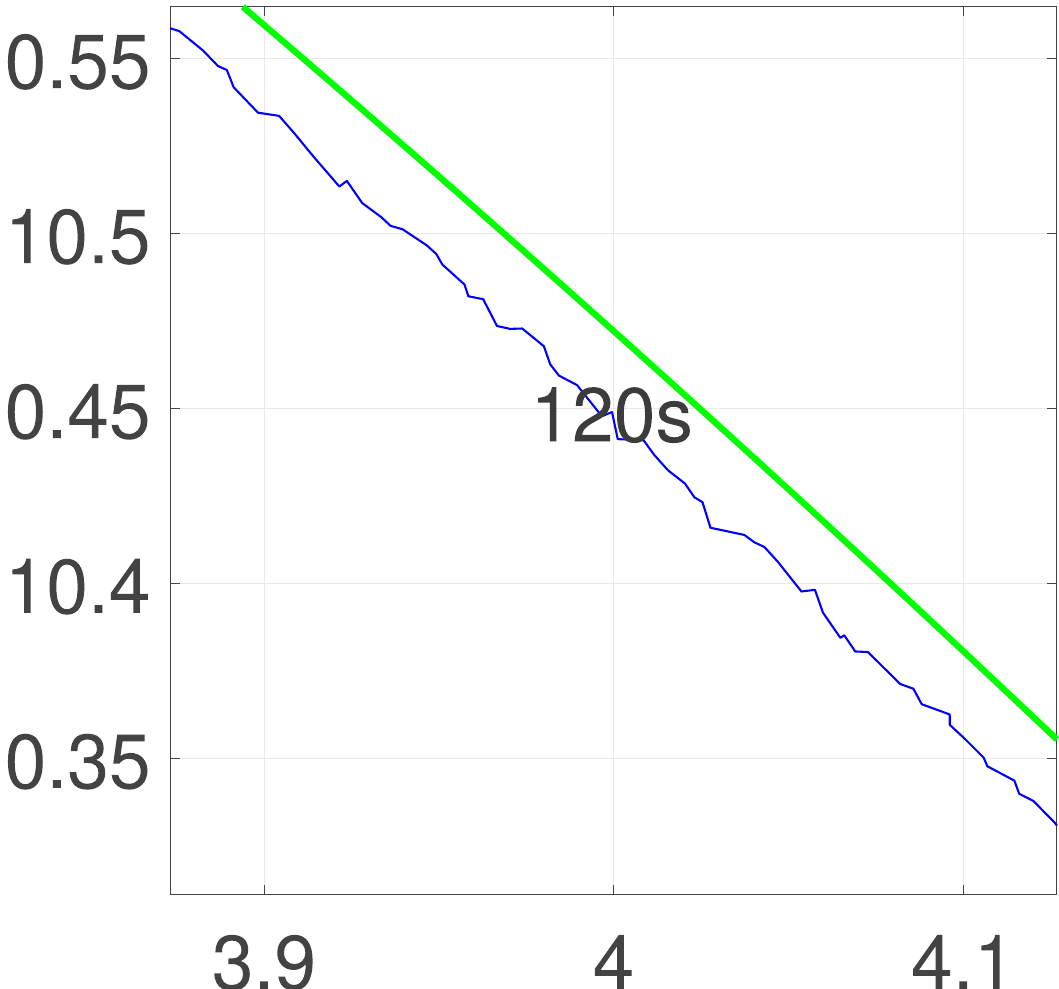}}
\end{overpic}
	\caption{{\bf Circumference trajectory.} A 12 [$m$] diameter circumference is used as trajectory reference. The small square shows a close up where error is appreciated, green line is the reference and blue one is the measured from VSLAM. The sequence in time is followed by the labels of seconds.}
	\label{fig:circulo}
\end{figure}
 \begin{figure}[t]
 	\centering
 	\begin{minipage}[t]{0.5\textwidth}
 		\includegraphics[trim={2.65cm 0.0cm 3.15cm 0.2cm},clip,width=0.95\textwidth]{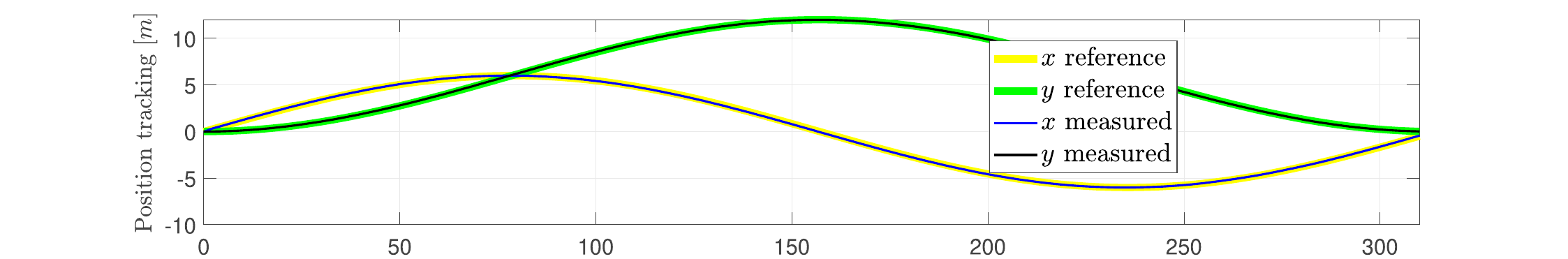}
 	\end{minipage}
 	
 	\centering
 	\begin{minipage}[t]{0.5\textwidth}
 		\includegraphics[trim={2.65cm 0.0cm 3.15cm 0.1cm},clip,width=0.95\textwidth]{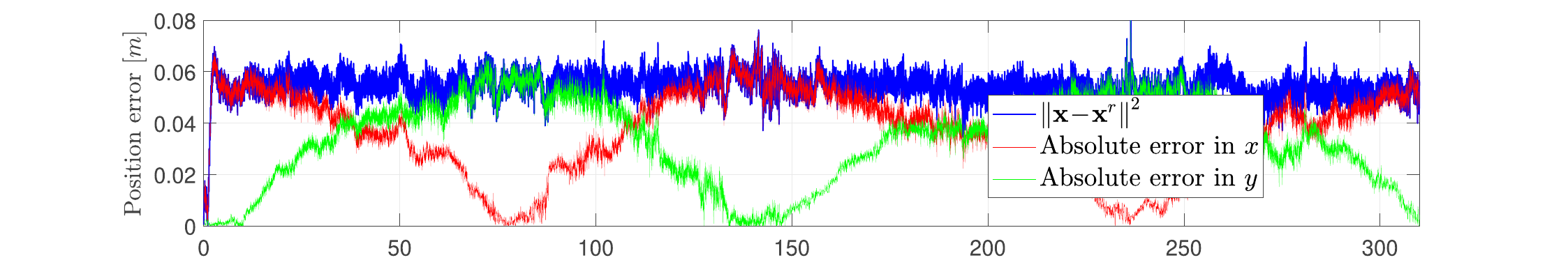}
 	\end{minipage}
 	
 	\centering
 	\begin{minipage}[t]{0.5\textwidth}
 		\includegraphics[trim={2.65cm 0.0cm 3.15cm 0.1cm},clip,width=0.95\textwidth]{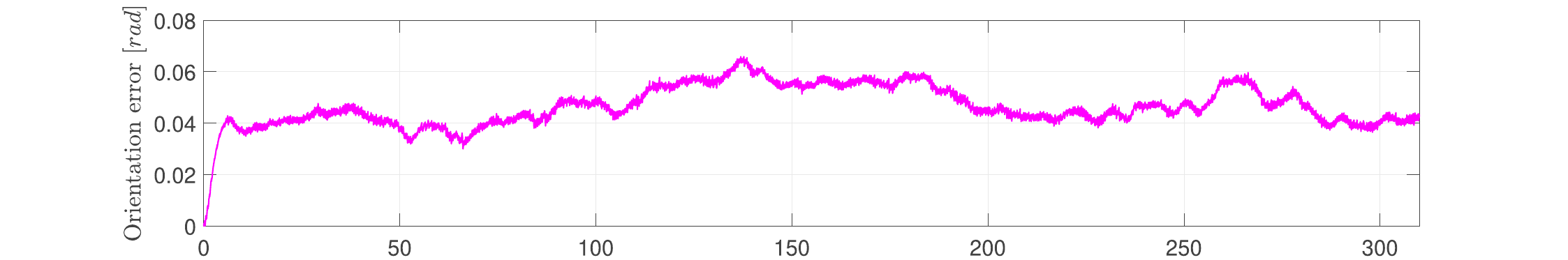}
 	\end{minipage}

 	\begin{minipage}[t]{0.5\textwidth}
 		\includegraphics[trim={2.65cm 0.0cm 3.15cm 0.1cm},clip,width=0.95\textwidth]{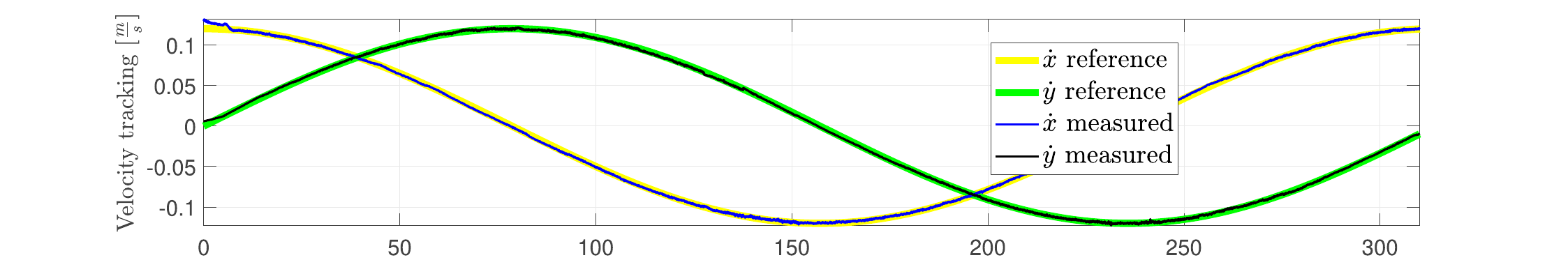}
 	\end{minipage}

 	\begin{minipage}[t]{0.5\textwidth}
 		\includegraphics[trim={2.65cm 0.0cm 3.15cm 0.1cm},clip,width=0.95\textwidth]{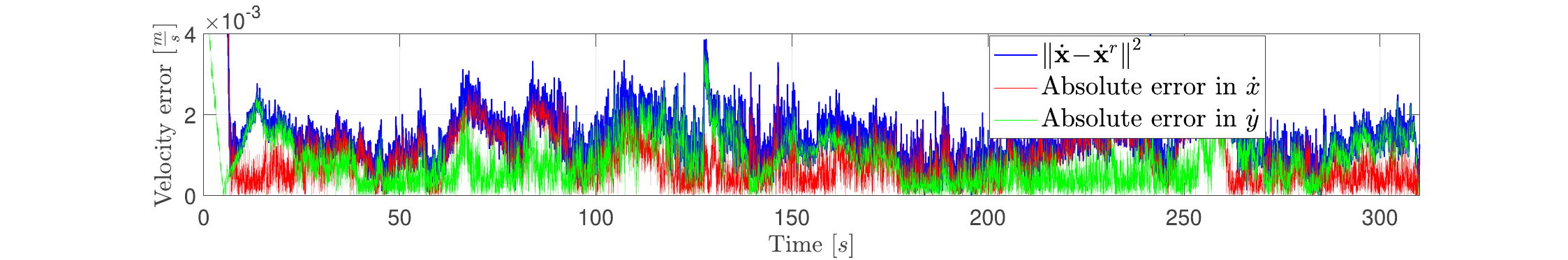}
 	\end{minipage}
 	\caption{{\footnotesize Circumference results during time. From top to bottom: position tracking comparison, position errors in $x$, $y$ and Euclidean norm; error in orientation, velocity tracking comparison and velocity errors in $x$, $y$ and Euclidean norm.}}
 	\label{fig:circulo_graphs}
 \end{figure}
The first trajectory to test is a 12-meter-diameter circumference with constant speed, see Figs. \ref{fig:overview} and \ref{fig:circulo}. The real-time NMPC signals are analyzed in Fig. \ref{fig:circulo_graphs} where the two upper graphs depict the tracking in position with an average error ($e_{{\scriptsize \mbox{AVG}}}$) of 5.42 [$cm$] and RMS error ($e_{{\scriptsize \mbox{RMS}}}$) of 5.45 [$cm$]. These errors are shown in Table \ref{table:errors} alongside with velocity errors.
	\setlength{\tabcolsep}{3pt}
	\begin{table}[t]	
		{\footnotesize
		\begin{center}
			\centering 
			\begin{tabular}{c|c|c|c|c}    \toprule
				 \textbf{Trajectory}$\rightarrow$ & \textbf{Circumference} & \textbf{Lemniscata} & \textbf{Multi Lemniscata} \\ \midrule
				\textbf{Position} $e_{{\scriptsize \mbox{AVG}}}$ & \textbf{5.42} [$cm$] & \textbf{4.97} [$cm$] & \textbf{5.00} [$cm$] \\ \midrule
				\textbf{Position} $e_{{\scriptsize \mbox{RMS}}}$ & \textbf{5.45} [$cm$] & \textbf{5.93} [$cm$] & \textbf{6.20} [$cm$] \\ \midrule
				\textbf{Velocity} $e_{{\scriptsize \mbox{AVG}}}$ & \textbf{1.4} [$mm/s$] & \textbf{3.5} [$mm/s$] & \textbf{3.8} [$mm/s$] \\ \midrule
				\textbf{Velocity} $e_{{\scriptsize \mbox{RMS}}}$ & \textbf{1.8} [$mm/s$] & \textbf{4.2} [$mm/s$] & \textbf{5.0} [$mm/s$] \\ \bottomrule
			\end{tabular}
		\caption{ Report of position and velocity errors.}
		\label{table:errors}
		\end{center} }
	\end{table}

 \begin{figure}[t]
 	\centering
 	\begin{minipage}{0.45\textwidth}
 		\includegraphics[trim={0.0cm 0.0cm 0.0cm 0.0cm},clip,width=\textwidth]{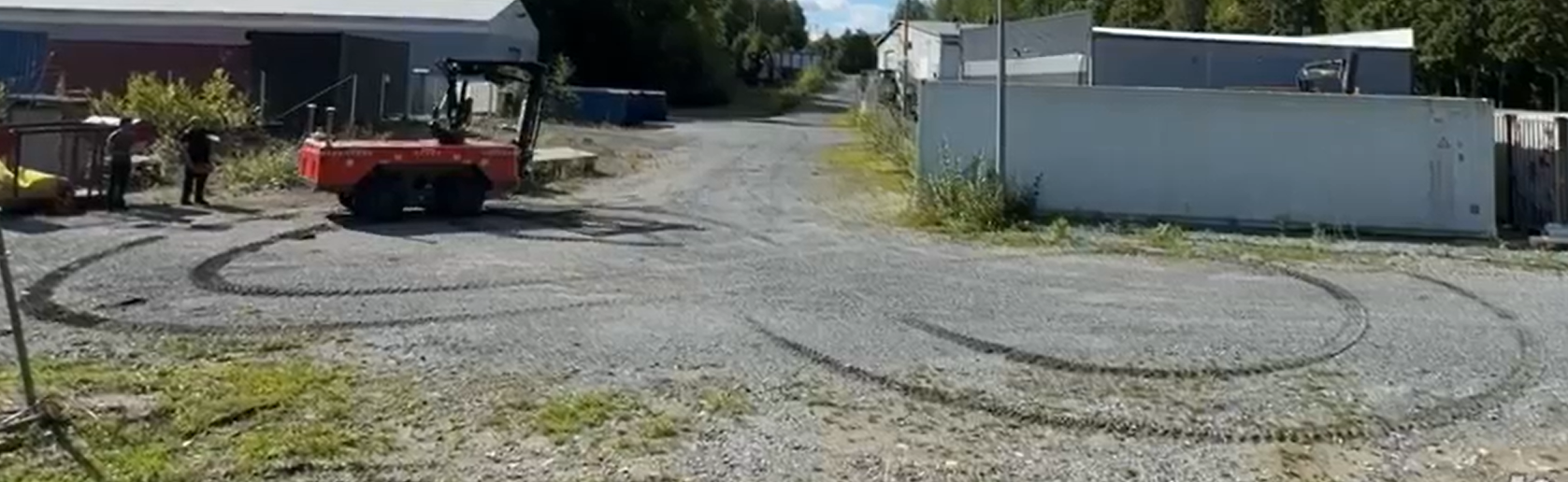}
 	\end{minipage}
 	
 	\centering
 	\begin{overpic}[width=0.45\textwidth,trim={2cm 0.1cm 3.6cm 1.6cm},clip]{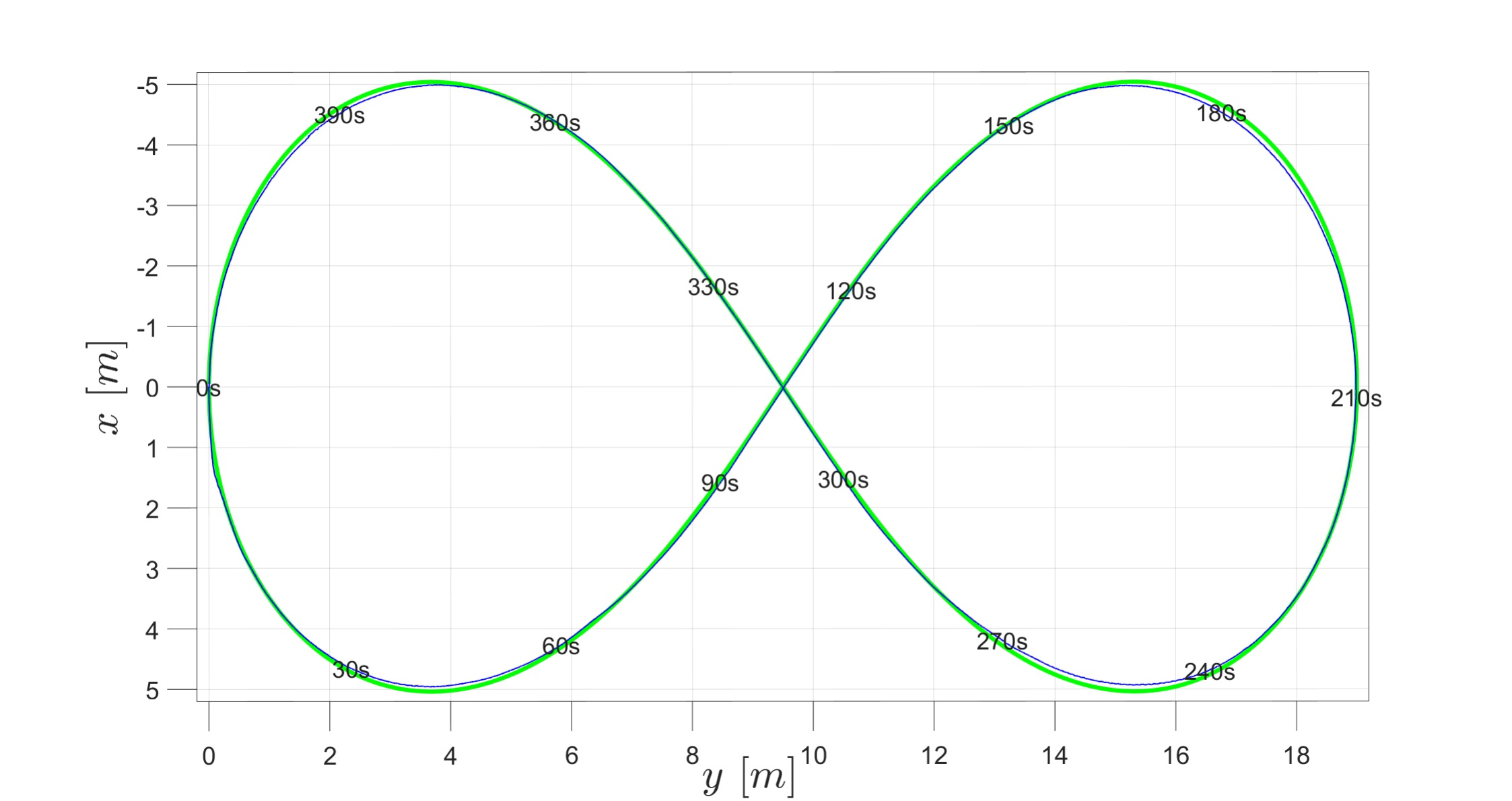}
    \put(67,19){\includegraphics[width=2cm,trim={9.3cm 1.5cm 10.8cm 1.4cm},clip]{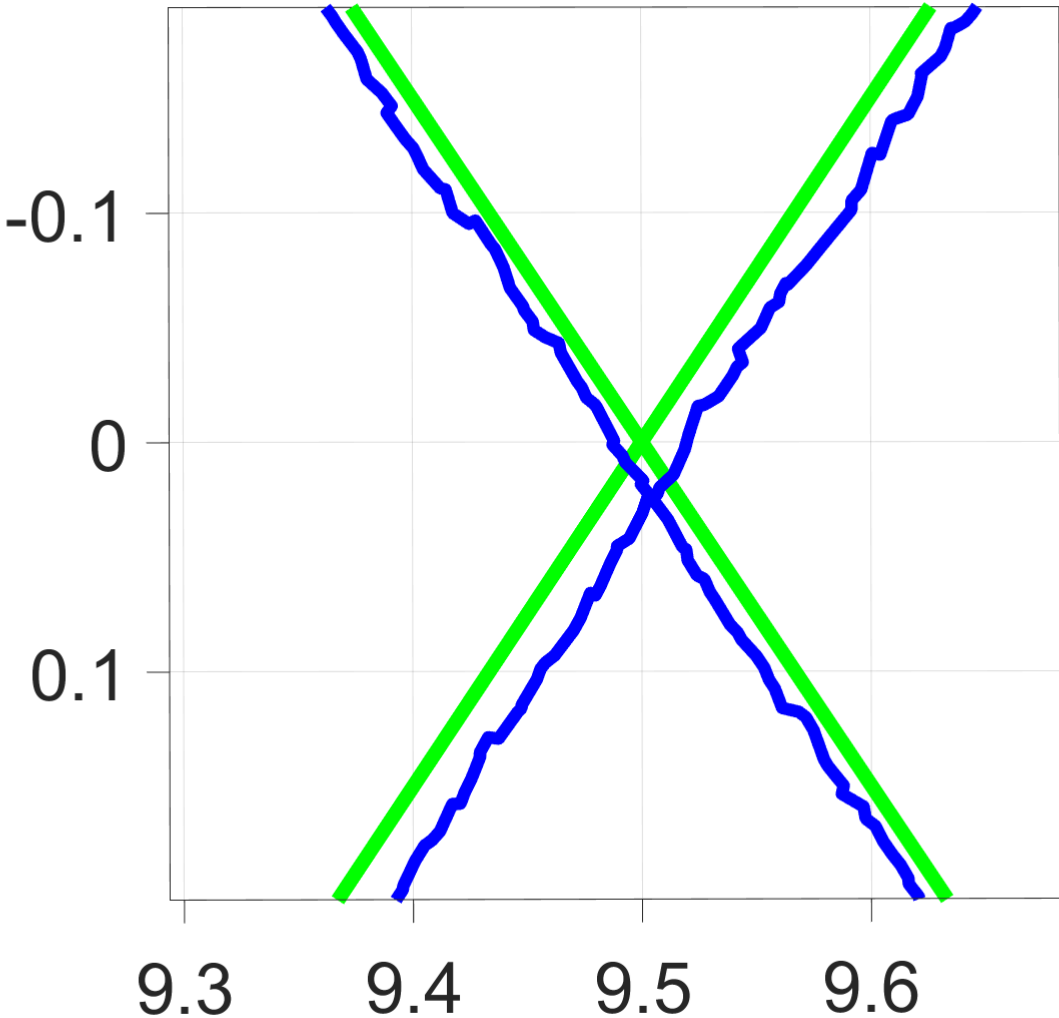}}
\end{overpic}
 	\caption{{\bf Lemniscate trajectory.} A 19$\times$10 [$m$] Lemniscate function is used as trajectory reference. The small square shows a close up where error is appreciated, green line is the reference and blue one is the measured from VSLAM. The sequence in time is followed by the labels of seconds.}
	\label{fig:lemnis_una}
 \end{figure}

 \begin{figure}[t]
 	\centering
 	\begin{minipage}{0.5\textwidth}
 		\includegraphics[trim={2.65cm 0.0cm 3.15cm 0.2cm},clip,width=0.925\textwidth]{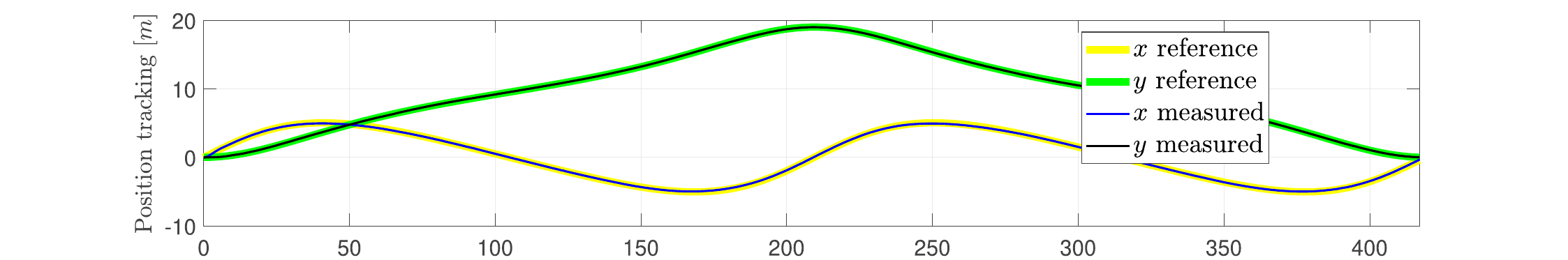}
 	\end{minipage}
 	
 	\centering
 	\begin{minipage}{0.5\textwidth}
 		\includegraphics[trim={2.65cm 0.0cm 3.15cm 0.1cm},clip,width=0.925\textwidth]{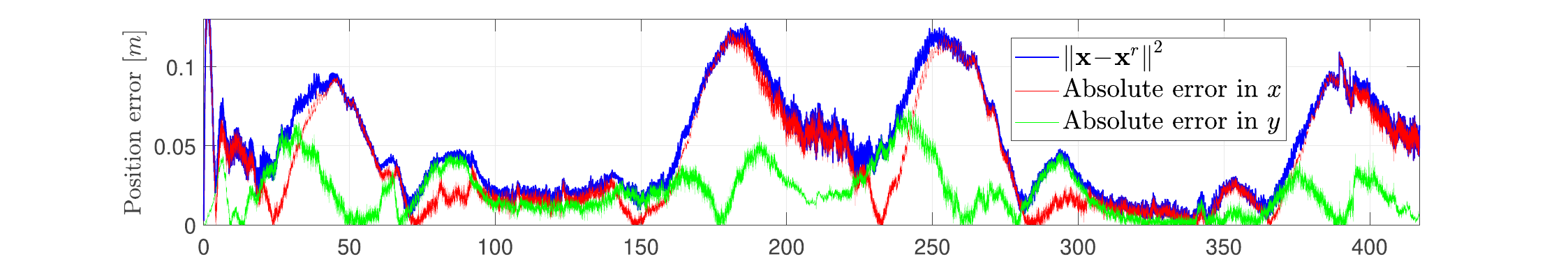}
 	\end{minipage}
 	
 	\centering
 	\begin{minipage}{0.5\textwidth}
 		\includegraphics[trim={2.65cm 0.0cm 3.15cm 0.1cm},clip,width=0.925\textwidth]{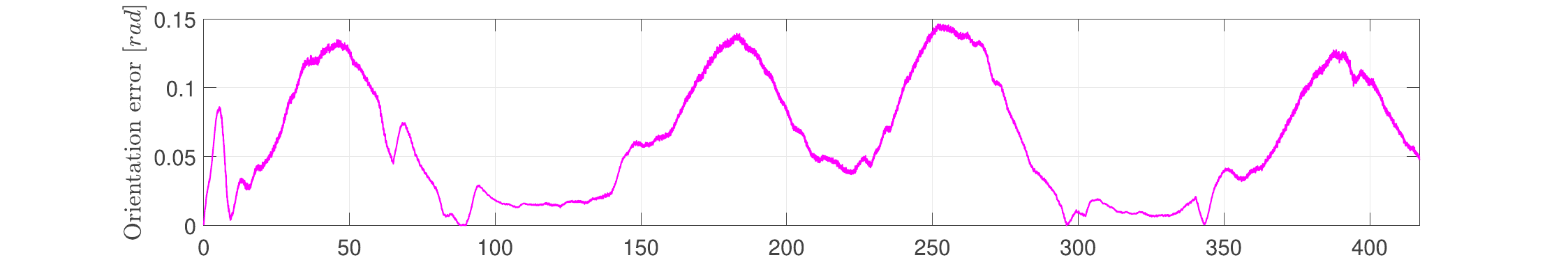}
 	\end{minipage}

 	\begin{minipage}{0.5\textwidth}
 		\includegraphics[trim={2.65cm 0.0cm 3.15cm 0.1cm},clip,width=0.925\textwidth]{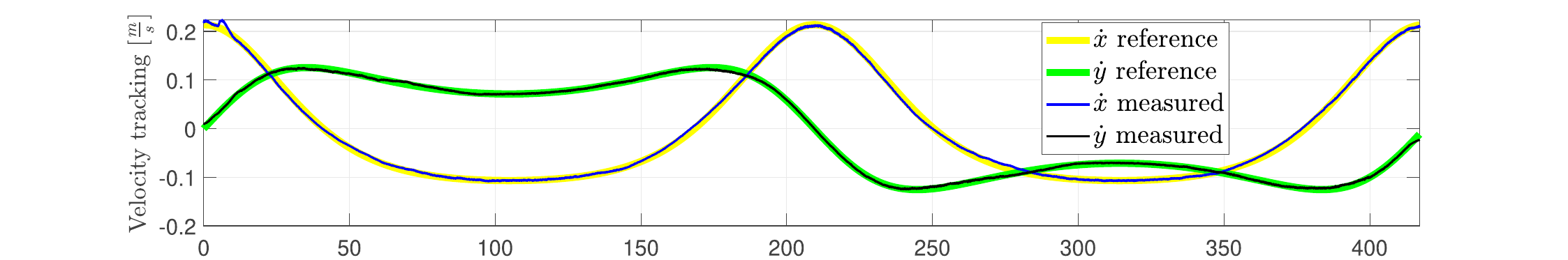}
 	\end{minipage}

 	\begin{minipage}{0.5\textwidth}
 		\includegraphics[trim={2.65cm 0.0cm 3.15cm 0.1cm},clip,width=0.925\textwidth]{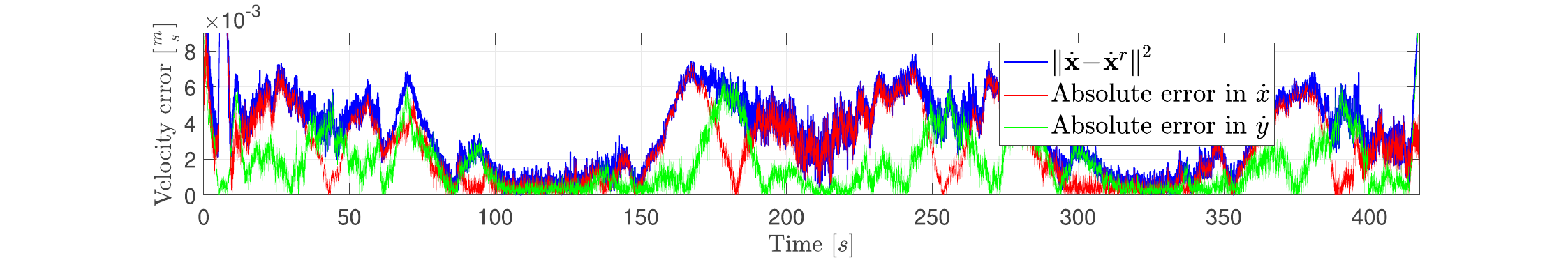}
 	\end{minipage}
 	\caption{Lemniscate results during time. From top to bottom: position tracking comparison, position errors in $x$, $y$ and Euclidean norm; error in orientation, velocity tracking comparison and velocity errors in $x$, $y$ and Euclidean norm.}
 	\label{fig:lemnis_una_graphs}
 \end{figure}
\begin{figure}[t] 
	\centering
	\includegraphics[trim={2.2cm 0.8cm 3.6cm 2.2cm},clip,width=6cm]{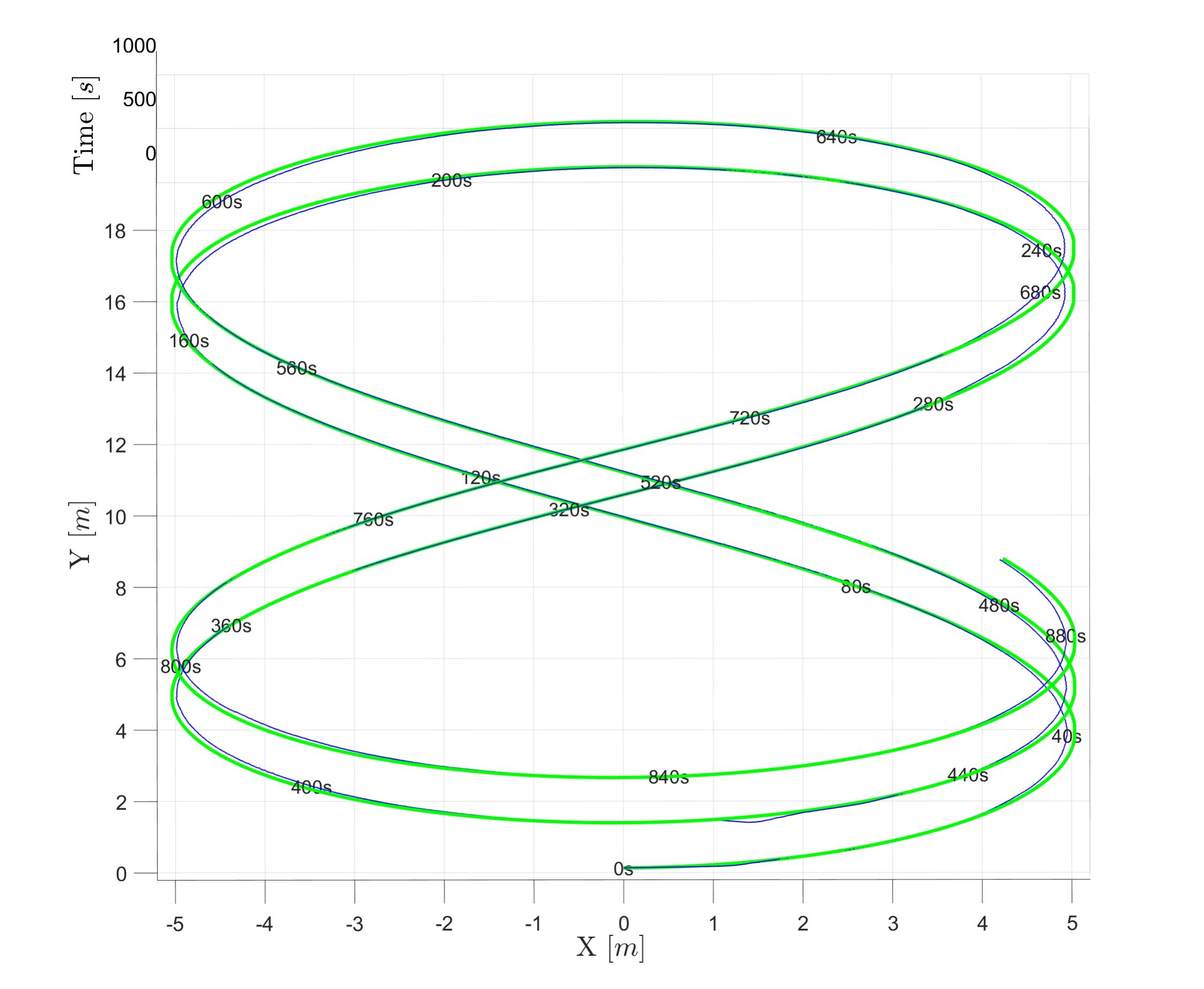}
	\caption{{\bf Multiple-Lemniscate trajectory.} The 19$\times$10 [$m$] Lemniscate function is used as trajectory reference for multiple iterations for long-life test of the NMPC. Green line is the reference and blue one is the measured from VSLAM. The sequence in time is followed by the labels of seconds.}
	\label{fig:88_doble}
\end{figure}
 \begin{figure}[t]
 	\centering
 	\begin{minipage}{0.5\textwidth}
 		\includegraphics[trim={2.65cm 0.0cm 3.15cm 0.2cm},clip,width=0.925\textwidth]{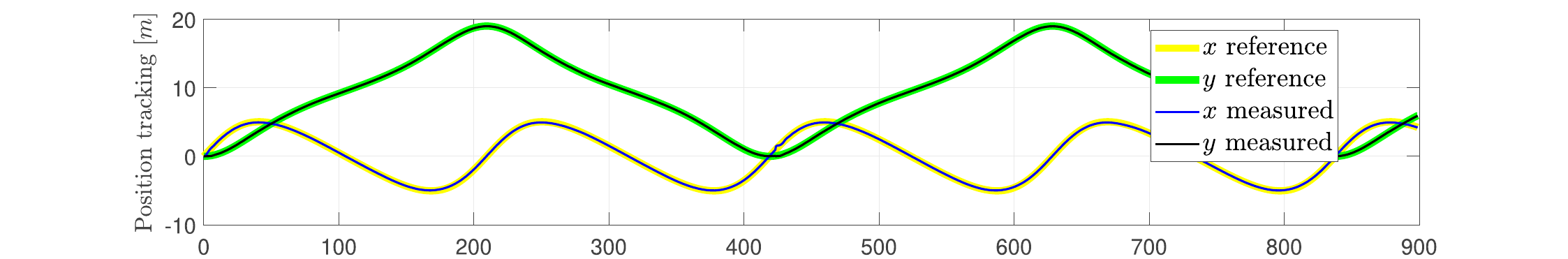}
 	\end{minipage}
 	
 	\centering
 	\begin{minipage}{0.5\textwidth}
 		\includegraphics[trim={2.65cm 0.0cm 3.15cm 0.1cm},clip,width=0.925\textwidth]{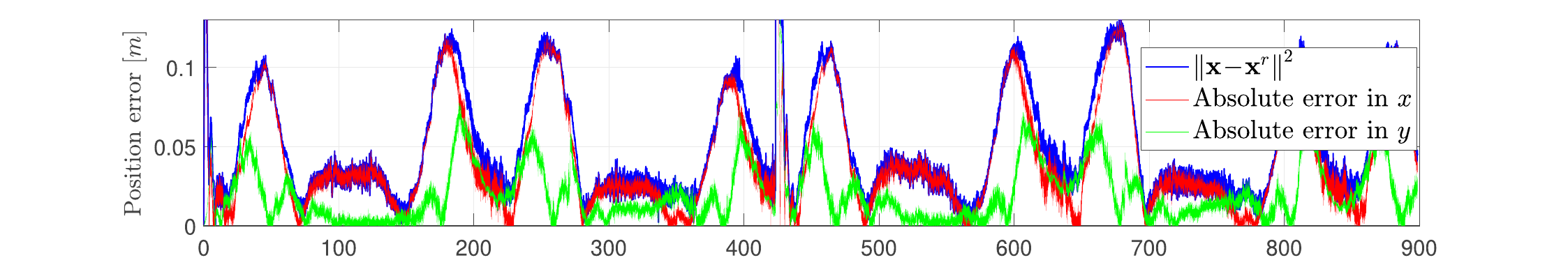}
 	\end{minipage}
 	
 	\centering
 	\begin{minipage}{0.5\textwidth}
 		\includegraphics[trim={2.65cm 0.0cm 3.15cm 0.1cm},clip,width=0.925\textwidth]{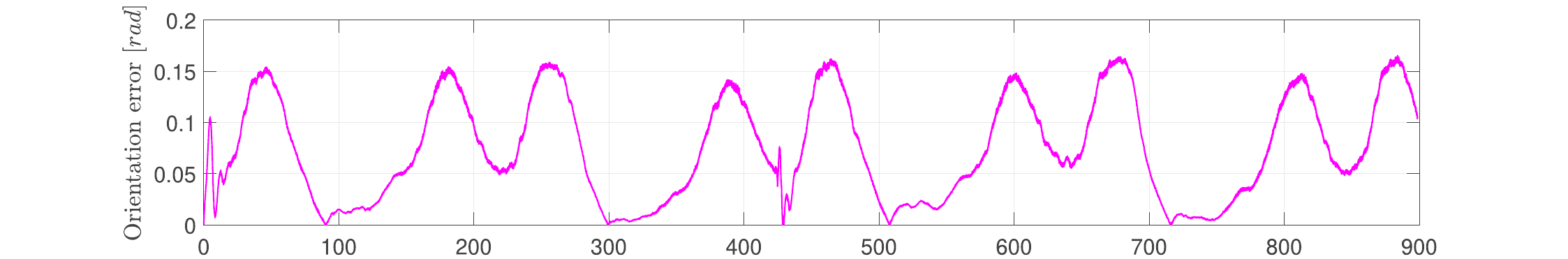}
 	\end{minipage}

 	\begin{minipage}{0.5\textwidth}
 		\includegraphics[trim={2.65cm 0.0cm 3.15cm 0.1cm},clip,width=0.925\textwidth]{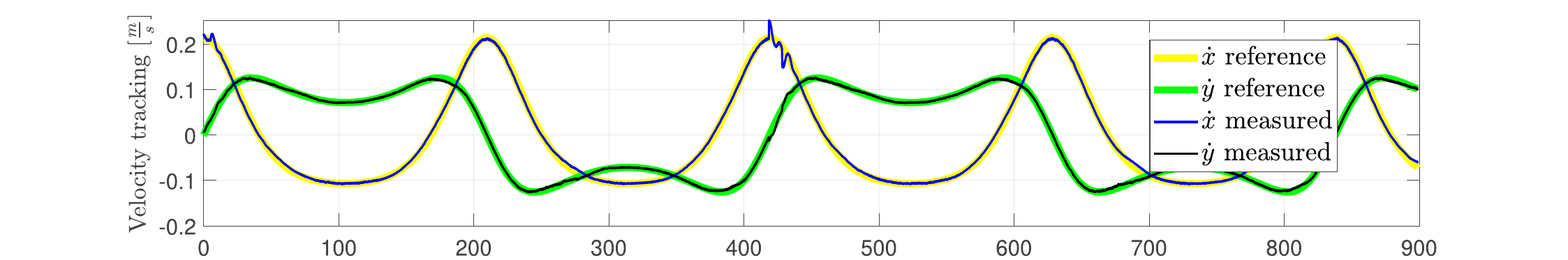}
 	\end{minipage}

 	\begin{minipage}{0.5\textwidth}
 		\includegraphics[trim={2.65cm 0.0cm 3.15cm 0.1cm},clip,width=0.925\textwidth]{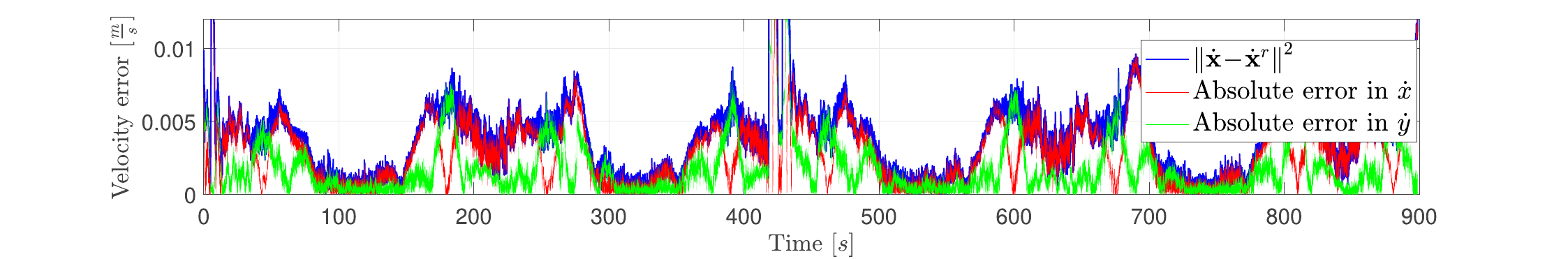}
 	\end{minipage}
 	\caption{{\footnotesize Multiple Lemniscates results during time. From top to bottom: position tracking comparison, position errors in $x$, $y$ and Euclidean norm; error in orientation, velocity tracking comparison and velocity errors in $x$, $y$ and Euclidean norm.}}
 	\label{fig:88_doble_graphs}
 \end{figure}

Additionally, we test a 19$\times$10 [$m$] Lemniscate trajectory, its shape is shown in Fig. \ref{fig:lemnis_una} while Fig. \ref{fig:lemnis_una_graphs} depicts the tracking results over time. The errors in position and velocity are also include in Table \ref{table:errors}. In order to test the long-life performance of our controller, we also run a set of multiple Lemniscates, see Figs. \ref{fig:88_doble} and \ref{fig:88_doble_graphs}.

According to Table \ref{table:errors}, the position errors oscillate, among these three trajectories, between 4.97 and 6.20 [$cm$]. Also, the velocity errors are between 1.4 and 5.0 [$mm/s$].

\section{Conclusions}
\label{sec:conclusion}
We have developed a novel real-time NMPC framework, capable of being accurate and fast simultaneously. According to Fig. \ref{fig:RT_sample} our NMPC is able to compute the current optimal solution around the same time that our fastest sensor does. Also, by managing the UTP sensor buffers in an appropriate manner we have avoided extra computational burdens, thus our algorithm fulfills the real-time requirements. The computational time is deterministic and bounded by the combination of warm start, high sensor rates and low maximum number of iterations of the solver; making our algorithm running in uniform execution times. Then, we have presented the accuracy performance of our NMPC with three different trajectories where the position and velocity errors are dramatically reduced (see Table \ref{table:errors}) with respect to approaches in the literature for similar trajectory shapes and sizes. The performance improvements of our NMPC is more noticeable when considering that it is implemented on a large heavy-duty mobile platform.

%









\bibliography{root,MD_Ref}

\begin{thebibliography}{10}

\bibitem{raj2022comprehensive}
R.~Raj and A.~Kos, ``A comprehensive study of mobile robot: history, developments, applications, and future research perspectives,'' {\em Applied Sciences}, vol.~12, no.~14, p.~6951, 2022.

\bibitem{rubio2019review}
F.~Rubio, F.~Valero, and C.~Llopis-Albert, ``A review of mobile robots: Concepts, methods, theoretical framework, and applications,'' {\em International Journal of Advanced Robotic Systems}, vol.~16, no.~2, p.~1729881419839596, 2019.

\bibitem{khan2021comprehensive}
R.~Khan, F.~M. Malik, A.~Raza, and N.~Mazhar, ``Comprehensive study of skid-steer wheeled mobile robots: development and challenges,'' {\em Industrial Robot: the international journal of robotics research and application}, vol.~48, no.~1, pp.~142--156, 2021.

\bibitem{diehl2009efficient}
M.~Diehl, H.~J. Ferreau, and N.~Haverbeke, ``Efficient numerical methods for nonlinear mpc and moving horizon estimation,'' {\em Nonlinear model predictive control: towards new challenging applications}, pp.~391--417, 2009.

\bibitem{bock1984multiple}
H.~G. Bock and K.-J. Plitt, ``A multiple shooting algorithm for direct solution of optimal control problems,'' {\em IFAC Proceedings Volumes}, vol.~17, no.~2, pp.~1603--1608, 1984.

\bibitem{diehl2006fast}
M.~Diehl, H.~G. Bock, H.~Diedam, and P.-B. Wieber, ``Fast direct multiple shooting algorithms for optimal robot control,'' {\em Fast motions in biomechanics and robotics: optimization and feedback control}, pp.~65--93, 2006.

\bibitem{fankhauser2018robust}
P.~Fankhauser, M.~Bjelonic, C.~D. Bellicoso, T.~Miki, and M.~Hutter, ``Robust rough-terrain locomotion with a quadrupedal robot,'' in {\em 2018 IEEE International Conference on Robotics and Automation (ICRA)}, pp.~5761--5768, IEEE, 2018.

\bibitem{grandia2023perceptive}
R.~Grandia, F.~Jenelten, S.~Yang, F.~Farshidian, and M.~Hutter, ``Perceptive locomotion through nonlinear model-predictive control,'' {\em IEEE Transactions on Robotics}, vol.~39, no.~5, pp.~3402--3421, 2023.

\bibitem{prado2020adaptive}
A.~J. Prado, D.~Ch{\'a}vez, O.~Camacho, M.~Torres-Torriti, and F.~A. Cheein, ``Adaptive nonlinear mpc for efficient trajectory tracking applied to autonomous mining skid-steer mobile robots,'' in {\em 2020 IEEE ANDESCON}, pp.~1--6, IEEE, 2020.

\bibitem{aro2023nonlinear}
K.~Aro, R.~Urvina, N.~N. Deniz, O.~Menendez, J.~Iqbal, and A.~Prado, ``A nonlinear model predictive controller for trajectory planning of skid-steer mobile robots in agricultural environments,'' in {\em 2023 IEEE Conference on AgriFood Electronics (CAFE)}, pp.~65--69, IEEE, 2023.

\bibitem{wang2023trajectory}
J.~Wang, Z.~Liu, H.~Chen, Y.~Zhang, D.~Zhang, and C.~Peng, ``Trajectory tracking control of a skid-steer mobile robot based on nonlinear model predictive control with a hydraulic motor velocity mapping,'' {\em Applied Sciences}, vol.~14, no.~1, p.~122, 2023.

\bibitem{aro2024robust}
K.~Aro, L.~Guevara, M.~Torres-Torriti, F.~Torres, and A.~Prado, ``Robust nonlinear model predictive control for the trajectory tracking of skid-steer mobile manipulators with wheel--ground interactions,'' {\em Robotics}, vol.~13, no.~12, p.~171, 2024.

\bibitem{Bib:Betts}
J.-T. Betts, {\em Practical Methods for Optimal Control and Estimation Using Nonlinear Programming}.
\newblock Advances in Design and Control, SIAM, 2nd~ed., 2010.

\bibitem{caracciolo1999trajectory}
L.~Caracciolo, A.~De~Luca, and S.~Iannitti, ``Trajectory tracking control of a four-wheel differentially driven mobile robot,'' in {\em Proceedings 1999 IEEE international conference on robotics and automation (Cat. No. 99CH36288C)}, vol.~4, pp.~2632--2638, IEEE, 1999.

\bibitem{kozlowski2004modeling}
K.~Koz{\l}owski and D.~Pazderski, ``Modeling and control of a 4-wheel skid-steering mobile robot,'' {\em International journal of applied mathematics and computer science}, vol.~14, no.~4, pp.~477--496, 2004.

\bibitem{Bib:murray}
R.-M. Murray, Z.-X. Li, and S.-S. Sastry, {\em A Mathematical Introduction to Robotic Manipulation}.
\newblock CRC Press, Boca Raton, FL, 1994.

\bibitem{shahna2025anti}
M.~H. Shahna, P.~Mustalahti, and J.~Mattila, ``Anti-slip ai-driven model-free control with global exponential stability in skid-steering robots,'' {\em arXiv preprint arXiv:2504.08831}, 2025.

\bibitem{khalil2002nonlinear}
H.~K. Khalil and J.~W. Grizzle, {\em Nonlinear systems}, vol.~3.
\newblock Prentice hall Upper Saddle River, NJ, 2002.

\bibitem{Bib:Nocedal}
J.~Nocedal and S.-J. Wright, {\em Numerical Optimization}.
\newblock Springer-Verlag, 1999.

\bibitem{mur2017orb}
R.~Mur-Artal and J.~D. Tard{\'o}s, ``Orb-slam2: An open-source slam system for monocular, stereo, and rgb-d cameras,'' {\em IEEE transactions on robotics}, vol.~33, no.~5, pp.~1255--1262, 2017.

\end{thebibliography}
\bibliographystyle{ieeetr}

\end{document}